\documentclass[10pt,journal]{IEEEtran}

\usepackage{float}
\usepackage{pifont}
\usepackage{geometry}
\usepackage[fleqn]{amsmath}
\usepackage{graphicx}
\usepackage{txfonts}
\usepackage{lineno,hyperref}
\usepackage{cite}

\newcommand\AI{\textit{AI}}

\modulolinenumbers[5]

\usepackage{xargs}                      
\usepackage[colorinlistoftodos,prependcaption,textsize=tiny]{todonotes}
\newcommandx{\unsure}[2][1=]{\todo[linecolor=red,backgroundcolor=red!25,bordercolor=red,#1]{#2}}
\newcommandx{\change}[2][1=]{\todo[linecolor=blue,backgroundcolor=blue!25,bordercolor=blue,#1]{#2}}
\newcommandx{\info}[2][1=]{\todo[linecolor=OliveGreen,backgroundcolor=OliveGreen!25,bordercolor=OliveGreen,#1]{#2}}
\newcommandx{\improvement}[2][1=]{\todo[linecolor=Plum,backgroundcolor=Plum!25,bordercolor=Plum,#1]{#2}}
\newcommandx{\thiswillnotshow}[2][1=]{\todo[disable,#1]{#2}}

\begin{document}

\title{Narrow artificial intelligence with machine learning for real time estimation of a mobile agent's location using Hidden Markov Models}

\author{
\IEEEauthorblockN{{ C\'edric Beaulac and Fabrice Larribe}\\} 
\and
\IEEEauthorblockA{D\'epartement de math\'ematiques, Universit\'e du Qu\'ebec \`a Montr\'eal \\ 201, avenue du Pr\'esident-Kennedy, Montr\'eal, Qu\'ebec, Canada}

}


\maketitle

\begin{abstract}We propose to use a supervised machine learning technique to track the location of a mobile agent in real time. Hidden Markov Models are used to build artificial intelligence that estimates the unknown position of a mobile target moving in a defined environment.  This narrow artificial intelligence performs two distinct tasks. First, it provides real-time estimation of the mobile agent's position using the forward algorithm. Second, it uses the Baum-Welch algorithm as a statistical learning tool to gain knowledge of the mobile target. Finally, an experimental environment is proposed, namely a video game that we use to test our artificial intelligence. We present statistical and graphical results to illustrate the efficiency of our method.  
 
\end{abstract}

\begin{IEEEkeywords}
Hidden Markov Model, User tracking, State estimation, Statistical learning, Non player characters
\end{IEEEkeywords}


\section{Motivation}

In this paper we address the problem of locating or tracking down a mobile target; specifically, we are interested in estimating the unknown position of a mobile agent. To do so, we rely on two main sources of information: our knowledge of the agent's usual behaviour and the information obtained from the environment (e.g. through vision, hearing). We build a proper solution to this problem by creating a mathematical model, an associated algorithm, and finally programming this algorithm; the result is autonomous artificial intelligence (\textit{AI}).

\smallskip

A common problem in video game environments when multiple games are played in succession is that a human player will improve with each game played against the same opponent because he or she learns the opponent's strategies, while the typical video game's \AI\ would not adapt to its opponent.  Yannakakis \& Togelius \cite{Yannakakis15} identify core research areas within the field of \AI\ in games. According to the list of areas they identify, our problem is to both create a \textit{believable agent} and to provide it with \textit{behavior learning} abilities.

\smallskip

 We were inspired by Hladky and Bulitko \cite{Hladky09} to use a Hidden Markov Model (\textit{HMM}) to achieve this goal. In that context, the hidden state is the unknown position of the mobile agent, and our problem is one of state estimation. Our main idea is to use the Baum-Welch algorithm\cite{Baum70} as a machine learning tool that uses the \AI's observations to build up knowledge on a mobile agent over time. This knowledge is represented by the transition matrix that is used by our \AI\ to estimate the hidden state at any time. This matrix contains the various probabilities that the mobile agent moves from one position to another, i.e. from one state to another. By construction, this matrix is central in state estimation; thus, the better our \AI\ estimates the transition matrix, the more accurately it can predict the mobile target's movements. Our goal is to demonstrate that this learning algorithm can be efficient in that precise context. By using this methodology our \AI\ should adapt to and learn from various situations, which makes it suitable for various competitive video games.

\section{Related work}

Particle models were first used in the context of state estimation problems in video game environments by Bererton \cite{Bererton04}. His article describes the actual state of \AI\ in video games, defines precisely the major problems and demonstrates that particle filters are useful in order to solve these problems. This paper inspired a number of other papers that resort to particle filters for addressing similar problems. For example, Weber \& al \cite{Weber11} used a particle model with multiple layers in order to represent the state of a complex game (e.g. an \textit{RTS} game) and proved the feasibility of the estimation of this state.  Southey \& al \cite{Southey07} were among the first to use hidden Markov models in a video game \AI\ context. They proved that these models could be useful in tracking the movement of various units in an \textit{RTS}. Finally, Hladky and Bulitko \cite{Hladky09} decided to compare the efficiency of particle models with hidden Markov models in order to estimate the position of a moving agent and noticed that Markov models might produce more precise estimates in certain situations.  A limitation of Hladky's model was that it had no behavior learning abilities. 

\smallskip

More recently,  Stanescu \& Certicky\cite{Stanescu16} used \textit{Answer Set Programming} in order to predict an opponent's production in a \textit{RTS} video game. Their problem shares many similarities with ours; for example, they only obtain incomplete knowledge on the current situation and they are also trying to predict some elements of the current game state. In order to predict unit production, they generate every valid unit combination and pick the most probable one as their estimate. As we will see in section \ref{sec:3.2}, the recursive structure of the \textit{forward} algorithm will naturally do all of these steps more efficiently in order to predict the opponent's location.

\smallskip

Reinforcement learning (\textit{RL}) is also a very popular approach to build strong video game \textit{AI}. Lately, Wang \& Tan \cite{Wang15} and Emigh \& al. \cite{Emigh16} approached the wider problem of building artificial intelligence that can \textit{win} a video game base solely on \textit{RL}. Even though both of these articles and many others demonstrate the incredible potential of \AI\ using \textit{RL}, these models also have their fair share of weaknesses. When it comes to building \AI\ for a rather complex video game, both the extremely large state space and high dimensionality will cause the learning process to be very slow. The uses of \textit{RL} will also result in \AI\ that takes actions that maximize the expected reward without really understanding why. In a certain way, this \AI\ never truly understands the goal of the game nor the multiple distinct components that can cause a player to win or lose a game.  Wang \& Tang \cite{Wang15} tackles briefly these problems by creating two distinct learning processes: learning the appropriate behavior and focusing on optimal weapon selection.  
  
\smallskip

Our approach is to further segment \AI\ learning process in order to create a \textit{believable agent} and to reduce the potential dimensionality problems. For instance, for a \textit{FPS} video game, we would like our \AI\ to understand that it must aim well, establish a good strategy, estimate its opponent's location as accurately as possible and that these subtasks are independent of one another.  Because of the results of \cite{Hladky09}, we decided to approach the estimation of an opponent's location with hidden Markov models. The Markov assumption is very natural when it comes to modelling movements and, therefore, utilizing a Markov model would be more appropriate for solving this precise problem as opposed to other articles that are trying to solve larger problems. To further improve the actual models, we will use an Expectation-Maximisation algorithm in order to give the \AI\ the ability to build up it's own knowledge based on its experience. This article contains the key results that are explained in full detail in the associated thesis \cite{Beaulac15}.   

\section{Hidden Markov Model techniques and notations}

In the following section, we define a Hidden Markov Model, the notations we use and the algorithms related to that model. This model consists of a Markov process and an observation function. First, let us define $X_t$ as the Markov process with state space $S = \{ s_1, s_2, ..., s_n \}$. Because we will work with discrete-time Markov processes, it is natural to define a transition function $a_{ij} = P( x_t = s_j | x_{t-1} = s_i)$, and an associated transition matrix \textbf{A}. Finally, the Markov process also consists of an initial distribution function $ \mu_i = P( x_1 = s_i )$ which is frequently represented as a vector $ {\bf\mu} = \{ \mu_1, \mu_2, ..., \mu_n \}$.

\smallskip

As a hidden Markov process, the realizations of the Markov process $x_t$ are hidden, i.e. they are not observed. What we obtain instead, the observation, is a random variable $Y_t$ dependent on the hidden state. To complete the definition of an \textit{HMM}, we need an observation function: $b_i(y_t)= P(Y_t = y_t | x_t = s_i)$.

\smallskip 

The challenge is to use a sample of observations $ \{ y_1, y_2, ..., y_T \}$ to make a statistical inference on the hidden Markov process $X_t$. For example, it would be useful to be able to evaluate $P( x_t = s_i \mid y_1, y_2, ..., y_t) = P(y_1,y_2,...,y_t \cap x_t = s_i) / P(y_1,y_2,...,y_t ) $. To proceed, we use forward and backward values as defined by Rabiner \cite{Rabiner89}. These values will be used in inference and real-time estimation of the state. We adopt the classic definition of the forward values, i.e. $ \alpha_t(i) = P(y_1,y_2,...,y_t \cap x_t = s_i)$. The forward values can be calculated recursively: 

\begin{equation} \label{alpha} 
\begin{split}
   \alpha_{t+1}(j) &=  \sum_i^n  \alpha_t(i)\ a_{ij}\  b_j(x_{t+1}). \\
\end{split}
\end{equation}

 The initial values are $ \alpha_1(i) = \mu_ib_i(x_1)$ $ \forall i$. Backward values are denoted as usual, $ \beta_t(i) = P(y_{t+1},y_{t+2}, ..., y_T | x_t = s_i) $, and they can also be calculated recursively in a similar manner: 
 
 \begin{equation} \label{alpha} 
\begin{split}
   \beta_t(i) = \sum_{j=1}^n a_{ij}b_j(x_{t+1})\beta_{t+1}(j) \\
\end{split}
\end{equation}

 using the initial values $\beta_T(i) = 1$ $ \forall i $. 
 
 \smallskip
 
Finally, the \textit{Baum-Welch} algorithm is needed for parameter estimation. This \textit{EM (Expectation-Maximization)} algorithm, adapted for Hidden Markov Model estimation, is the best known and most widely used algorithm related to Hidden Markov Model inference. With this tool, every parameter of our model can be estimated. Note that in our context the initial distribution $\mu$ and the observation function $b$ will be known and, thus, need not be estimated. 

\smallskip

Forward and backward values are calculated as described earlier using the initial values submitted to the \textit{Baum-Welch} algorithm. We can use these values to compute various expected values, two of which are of particular interest. First:

\begin{equation} \label{gam}
\begin{split}
\gamma_t(i)  &=  P[x_t= S_i \mid y_1, y_2,...,y_T,]  \\ 
&= \alpha_{t}(i)\ \beta_t(i) / L_T, \\
\end{split}
\end{equation}

\noindent where $L_T$ is the likelihood of the observations. Using the forward and backward values we can also compute the following expected value:

\begin{equation} \label{xi}
\begin{split}
\xi_t(i,j) &=  P[x_{t-1}= s_i, x_{t} = s_j \mid y_1, y_2,...,y_T,]  \\ 
&= \alpha_{t-1}(i) a_{i,j} b_{j}(y_{t})\beta_{t}(j) / L_T. \\
\end{split}
\end{equation}

Finally, after we've computed $\gamma_t(i)$ and $\xi_t(i,j) $ for all $t,i$ and $j$, we can perform the maximization step. As mentioned, only the transition function has to be estimated. We have to find the transition matrix that maximizes the likelihood of our observations. We thus obtain an estimator very similar to the classic Markov model maximum-likelihood estimator:

\begin{equation} \label{aij}
\begin{split}
\hat{a}_{i,j} &= \frac{\sum_{t=2}^T \xi_t(i,j)}{ \sum_{t=2}^T \gamma_t(i)} = \frac{\sum_{t=2}^T \xi_t(i,j)}{ \sum_{t=2}^T \sum_j^n \xi_t(i,j)} \\
&= \frac{\sum_{t=2}^T  P [ x_{t-1} = s_i , x_t = s_j \mid y_1 , y_2 ,...,y_T ] }{ \sum_{t=2}^T P [ x_t = s_i \mid y_1 , y_2 ,...,y_T ] }.  
\end{split}
\end{equation}

\smallskip

To conclude this section, we present the details of how the algorithm works. The first step is to submit the initial values of the Hidden Markov Model's parameters to the algorithm. These consist of the initial distribution function, the transition function and the observation function. The second step is to calculate forward and backward values using the parameters we currently have. The third step involves computing the expected values as we did in equation (\ref{gam}) and (\ref{xi}) using the $\alpha$'s and $\beta$'s we computed in the previous step. Finally, the maximization step consists of using both of these expected values to estimate the parameters of the Hidden Markov Models. We then repeat this process starting from the second step until the parameter's estimates are stable.  

\smallskip

To summarize, here are the steps of the algorithm. 

\smallskip

1) Submit the initial values of the parameters \textbf{A}, $\mu$ and $b$.

\smallskip

2) Compute $\alpha_t(i)$ and $\beta_t(i)$, $\forall t$ and $\forall i$ using the forward and backward algorithms.

\smallskip

3) Compute the expected values $\gamma_t(i)$ and $\xi_t(i,j)$, $\forall t$ and $\forall i,j$ using the forward and backward values.

\smallskip

4) Estimate the parameters \textbf{A}, $\mu$ and $b$ using the expected values as in (\ref{aij}).

\smallskip

5) Return to step 2) and repeat until a desired level of convergence is attained.

\section{Main idea of our approach}

\subsection{Mathematical modelling}

First, we define what the Markov process represents. Recall that our main goal is to estimate the unknown position of a mobile agent in a restricted area; in our experimental environment, which is a competitive video game, the mobile agent we are tracking is our opponent. Because its position is unknown, we suppose that its movements follow a Markov chain, i.e. the location where the agent will be at time $t+1$ depends only on its location at time $t$. More formally, we define our Markov process, $X_t$, as the position of the mobile agent, at time $t$. Because this position is unknown we never observe the realizations of this process, but we obtain observations that are dependent on these hidden realizations. These observations are the realizations of a Hidden Markov Model. 		

\smallskip

Next, we define the numerous parameters we need to work with an \textit{HMM}. First, we need to define the state space of the Markov process. To this end we must introduce our defined environment, since we are trying to estimate the unknown position of a mobile agent in a known restricted area. In the context of a video game, we call this the map, which is the equivalent of a sports field; it is where the action takes places. For example, in the context of smart vehicle technologies the map could be a certain city. Our goal is to create a methodology that efficiently estimates a mobile agent's location on this map. To begin, we grid this map, creating possible positions, and the set of all these positions forms our finite state space $S = \{ s_1, s_2, ..., s_n \}$. 

\smallskip

Now we describe how we define our initial distribution, $\mu$. In our video games application, we will assume that the initial distribution is directly implemented in the game, and known by all players. As most competitive video games have fixed starting locations for each player, a comparison can be made to a game of chess where each pawn starts at a precise location. Some video games have a set of possible starting locations, one of which is chosen randomly. Generally, these possible locations are near one another and are known by every experienced player. Thus, the initial distribution is considered to be known by our \AI\ in our experimental set up.  

\smallskip

As mentioned earlier, Hidden Markov Models involve an unobservable realization of the underlying Markov process $x_t$, and an observation which is a random variable $Y_t$ dependent on the hidden realization. In other words, the observations represent the stream of information received from the environment. In our context, the observations are the set of positions we can eliminate at each unit of time and, thus, are random and dependent on the hidden state (the actual position of the mobile agent). The problem consists of using these observations, via a well-constructed observation function, in a manner that will help us track down the agent. In our experimental environment, our methodology is used by an \AI\ battling against a human player. In that context, the observation is the set of positions observed by the \AI's avatar inside the video game at each time frame.

\begin{figure}[H]  
\noindent
\begin{center}
	\includegraphics[width=7cm]{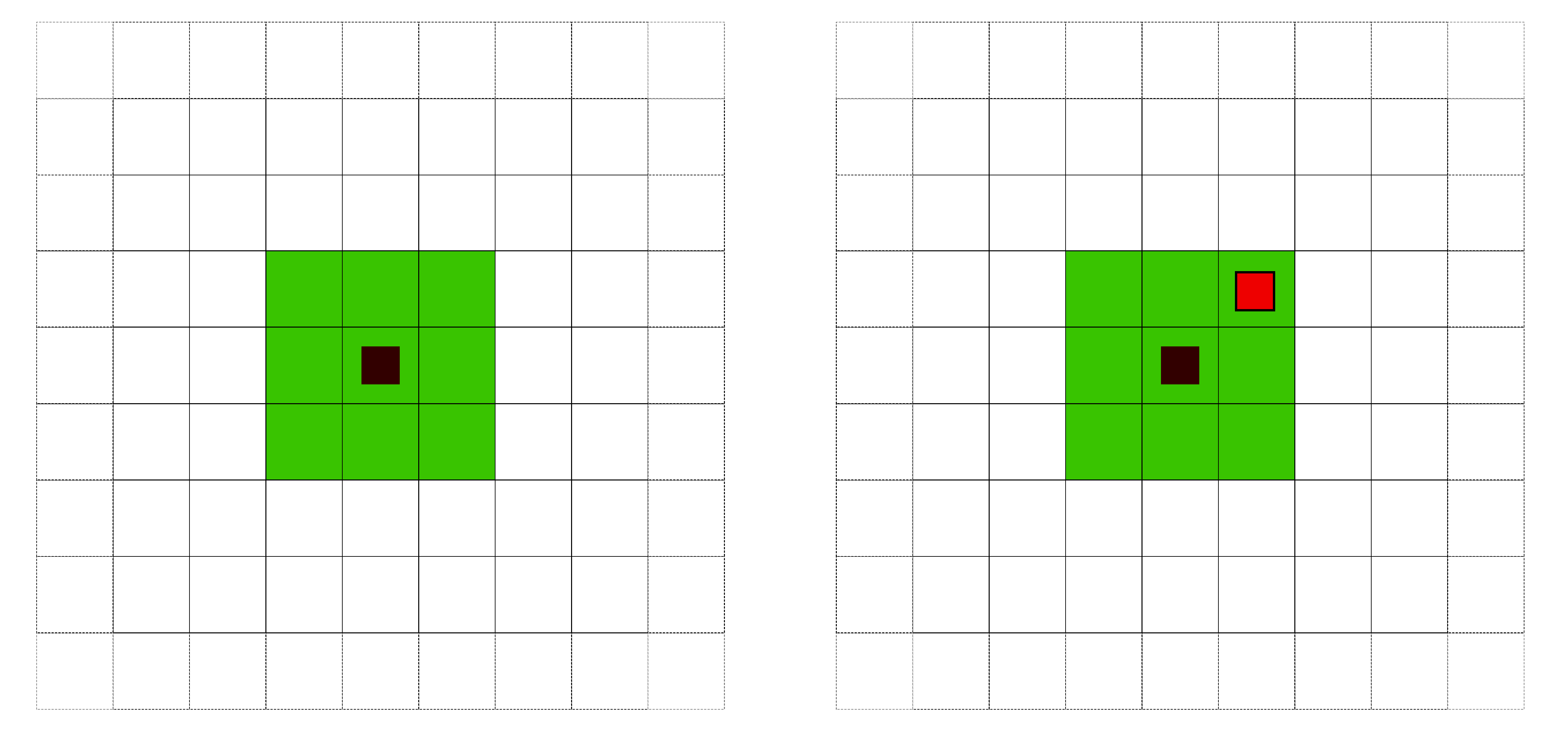} 
\caption{A representation of a vision radius, common in video games. The dark brown square represents our avatar, the red square the mobile agent, and the green tiles the positions we observe.}
\label{radius}
\end{center}
\end{figure}

Figure \ref{radius} depicts two situations faced when tracking down a mobile agent. We either directly observe the agent or not. If we see the mobile target, as in the right image in Figure \ref{radius}, the estimate of the current hidden state is quite simple: it is not an estimate, it is agent's actual position. The model gets interesting when we actually have to estimate our opponent's position, as in the left image in Figure  \ref{radius}. 

\smallskip 

Suppose we are in a situation where estimation is required, and let us define the observation function $b$. To proceed, we introduce $W$, which is the set of all visible positions, namely the green tiles in Figure \ref{radius}. In other contexts, these observations can be obtained with human eyes, camera, satellite or even other senses such as hearing.  At this precise time, these are the locations where we know the agent is not found. They form a set of states we can eliminate as the possible hidden state of the Markov process. What we observe is that the agent is not at these positions. We develop the following observation function:

\begin{equation} \label{eq:1}
b_i(y_t) = P( Y_t = y_t \mid X_t = s_i ) = 1 -  \bf{1}[s_i \in W].
\end{equation}

Note that we are in a situation where we observe a set of positions $W$ but we do not observe the mobile agent we are tracking. This simple but logical definition of $b$ implies that the probability of obtaining our observation, which is not seeing the agent in any elements of $W$, conditional on the fact that he is there is 0. It also implies that the probability of not seeing the agent in the set of locations we observed knowing he is not in this set is 1. This binary observation function not only cancels the set of visible locations but also eliminates every path that would have passed by these positions at this particular time, due to the recursive nature of the forward algorithm. We'll explain this in detail in the next subsection. Note that if the hidden state is $s_i$ at time $t$ and that at this very time, $ s_i \in W$, we simply define $b_i(y_t) = 1$ and $b_j(y_t) = 0$ $ \forall j \neq i$. 

\smallskip

Notice that this observation function could be modified easily in order to include uncertainties about certain position by defining a function that take values somewhere between 0 and 1 at some point. This could prove useful if we were interested in using other types of observations such as the sounds produced by the mobile agent, which are useful observations in multiple video games and other applications. 

\smallskip

To build the desired \AI\, we need to use a Hidden Markov Model for two distinct tasks, namely: real-time estimation of the mobile agent's position, and machine learning for improved future estimation. It should be noted that Hladky and Bultiko \cite{Hladky09} and Southey \& al. \cite{Southey07} used a hidden semi-Markov Model instead of a Hidden Markov Model in their research. Even though we considered using this model, we found that the gain in realism was not worth the programming cost because it directly affected the efficiency of the learning component of our \AI. The Baum-Welch algorithm for a Hidden semi-Markov Model was indeed slower to compute. Using such an algorithm in the context of a video game might create enormous loading time between games, and the gain of realism might not help the \AI\ that much. For the benefit of the machine learning part of the \AI\ we are designing, we decided to use a standard Hidden Markov Model.

\smallskip

\subsection{Real-time state estimation} \label{sec:3.2}

 Recall that if we observe the mobile agent, estimation is useless. Therefore, let us suppose we are in a situation where estimation is required, i.e. we de not observe the mobile agent. We evaluate the probability that the Markov process is in state $s_i$ at time $t$ considering current and past observations for all states at all discrete moments. This probability is denoted $P_t(s_i)$ and is calculated using the forward algorithm:

\begin{equation} \label{eq:2} 
\begin{split}
P_t(s_i) &= P( x_t = s_i \mid Y_1 = y_1, Y_2 = y_2, ..., Y_t = y_t ) \\
&= \frac{P( x_t = s_i \cap Y_1 = y_1, Y_2 = y_2, ..., Y_t = y_t )}{P( Y_1 = y_1, Y_2 = y_2, ..., Y_t = y_t )} \\
&= \frac{\alpha_t(i)}{ \sum_{i=1}^n  \alpha_t(i)}. \\
\end{split}
\end{equation}

 Let us suppose that  we observe the position $s_i$ at time $t$. Because of the way the observation function is set up, $b_i(y_t) = 0$ and thus $P_t(s_i) = 0$, which is desired and which contributes to eliminating a set of possible positions at time $t$. Specifically, this methodology eliminates every state $s_i \in W$ if the agent is not observed. 

\smallskip
 
That being said, the important part is the fact that knowing the hidden state was not $s_i$ at time $t$ implies not only that $P_t(s_i) = 0$ but also that $\alpha_t(i) = 0$.  Because the $\alpha_{t+1}$'s are computed recursively using the $\alpha_t$'s, it simplifies further calculations by eliminating elements of the summations. In fact, it eliminates every state at time $t+1$ that requires the agent to be at state $s_i$ at time $t$. Additionally, from our tracking point of view, it also eliminates every path that passes through any positions $s_i \in W $ for every turn. This is the main driver for this methodology's efficiency. 
 
\subsection{Machine learning}

Recall that a common problem in video game environments when multiple games happen consecutively on the same map is that a human player will improve with each game played against the same opponent because he or she learns the opponent's strategies, while the typical video game's \AI\ would not adapt to its opponent. We refer to this ability of human players as adaptive memory. In any context where multiple experiments take place in the same environment, the machine learning process we are using could aid the artificial intelligence's learning from each experiment. Given that observations only eliminate slowly some possible positions throughout the experiment,  real-time estimation relies heavily on the transition matrix \textbf{A}. It represents all the information about the mobile agent the \AI\ possesses. Our goal is to update this information after each set of observations is obtained. We choose to use the Baum-Welch \cite{Baum70} algorithm using multiple independent sets of observations to create the transition matrix that represents the agent's behaviour. We built a slightly modified version from the one presented in equation \ref{aij}. To build that equation, we assume each set of observations to be independent and we were inspired by the maximum likelihood estimator of a regular Markov model. At each iteration the transition matrix is re-estimated using this formula:

\begin{center} \label{maij}
\begin{equation}
{\hat{a}}_{i,j} = \frac{ \sum^m \sum_{t=2}^T \xi_t(i,j)}{\sum^m \sum_{t=2}^T \sum_j^n \xi_t(i,j)},
\end{equation}
\end{center}

\noindent where $m$ is the number of experiments realized.

\smallskip

Because we want our \AI\ to be flexible and to detect changes in strategies, we will not use every game in our machine learning phase. Let us say the \AI\ has played 100 games against a certain player. The \textit{Baum-Welch} algorithm will use all of these 100 observation sequences to estimate the transition matrix \textbf{A}. Therefore, if the human player changes his strategy drastically, it will take multiple games until this new strategy has enough weight in the estimation to change the estimate of the transition matrix. For this reason, the matrix used by the \AI\ for real-time state estimation will be a mix between a general long-term knowledge of the agent and short-term knowledge represented by a transition matrix built upon only the last few games. This matrix will be more detailed in the next section.

\section{Experimental set up and results} \label{secres}

After developing the algorithm and implementing it in $C$++ we were interested in testing this new \AI\ with adaptive memory. For this we had to design our own experimental environment, a video game, which was also implemented in $C$++. We designed a turn-based game in which the two players play one at a time and make one move per turn. The human player can see up to two tiles away, while the \AI\ only see the positions directly surrounding it, as seen in Figure \ref{radius}.

\smallskip

In the game we created for testing purposes, the goal of the human player is to get to a particular position and stay there for a known number of turns. In contrast, the goal of the \AI-controlled character is to prevent that from happening by \textit{touching} its opponent. We chose this specific objective because it was directly affected by the quality of the estimations and because it did not require that we code a complicated strategy surrounding the \AI's movement. We invented a pirate concept to intrigue the players who ran tests on our game. In this concept, the player is a pirate trying to steal a treasure from the captain on a remote island.

\smallskip

One game mechanic we added was parrots, to go with our pirate concept. They are scattered on various tiles on the map, and the human player knows where they are. They serve as ``cameras'' for the captain, as they constantly give the \AI\ observations to inform it whether the mobile agent is on one of these tiles. They not only help the \AI\ by giving it more observations, but they also motivate the player to build complex strategies that take the cameras into account.

\smallskip


\begin{figure}
\begin{center} 
\includegraphics[width=7.2cm]{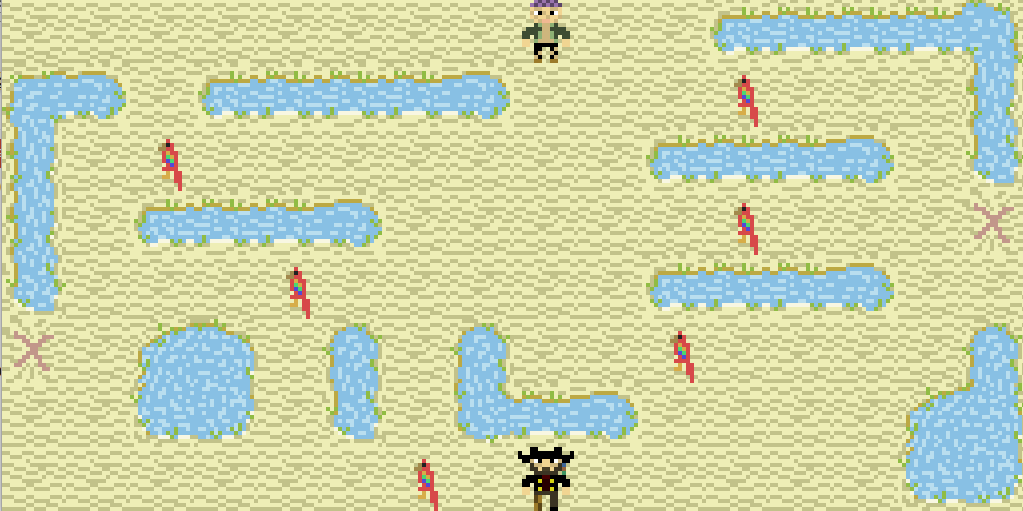}
\caption{An image of the video game we designed as an experimental environment. The player controls the humanoid on top, while the \AI\ controls the humanoid that starts at the bottom. Goals are marked by the red Xs on the map.}
\label{game}
\end{center}
\end{figure}

\smallskip

Figure \ref{game} represents the map we designed. In this figure, players are currently at their respective initial positions.

\smallskip

The \AI\  obtains observations from its environment as defined by equation \ref{eq:1}. At each turn, a vector of observations is created. All elements of this vector are 1's except the elements representing observed positions, which are 0's. The set of observed positions is the set of tiles in the vision radius of the \AI, as illustrated in Figure \ref{radius}, and the tiles observed by the cameras, corresponding to the parrots in our concept. These observations are used by the \AI\ along with the transition matrix, which represents information on the opponent's behavior, and the initial distribution to estimate the unknown position of its opponent as described in equation \ref{eq:2}.

\smallskip 

For this experiment, we also had to code a path-finding algorithm. Even though our primary goal was to estimate the position of a mobile agent, we needed the \AI\ to move autonomously in order to obtain data and conduct tests. Using equation \ref{eq:2}, the \AI\ can find the most probable hidden state and define it as its estimate of its opponent's position at each turn. This position is generally defined as the goal in the path-finding literature. Considering each state as a node in a graph, we implemented the Dijkstra's algorithm \cite{Dijkstra59} to let the \AI\ determine the shortest path to that estimate and start moving towards it. At each turn, the \AI\  determines a new goal, calculates the distance between the goal and each of its surrounding positions, and chooses a position to move from among these surrounding positions.

\smallskip

To prevent convergence problems, we used various logarithmic transformations, that were inspired by Rabiner \cite{Rabiner89} and that are well explained by both Durbin et al. \cite{Durbin98} and Beaulac \cite{Beaulac15}. 

\smallskip

 Finally, note that the forward and backward algorithms were coded using a matrix notation as presented by Zucchini \cite{Zucchini09} to simplify computations. Given that these values are a sum of a product, matrix multiplication can be useful. The forward vector $\alpha_t$, can be obtained by calculating the following: $\alpha_t = \mu  \mbox{\textbf{B}}(y_1) \mbox{\textbf{A}} \mbox{\textbf{B}}(y_2) \mbox{\textbf{A}} \cdot \cdot \cdot  \mbox{\textbf{A}} \mbox{\textbf{B}}(y_t)$ and the backward values: $\beta_t =  \mbox{\textbf{A}} \mbox{\textbf{B}}(y_{t+1}) \mbox{\textbf{A}} \mbox{\textbf{B}}(y_{t+2}) \cdot \cdot \cdot  \mbox{\textbf{A}} \mbox{\textbf{B}}(y_T)$ where : 

 \[
\textbf{B($y_t$)}
=
\begin{bmatrix}
    b_1(y_t) & 0 & 0 & \cdots & 0 \\
    0 & b_2(y_t) & 0 & \cdots & 0 \\
    \vdots & \vdots & \vdots & \ddots & \vdots \\
    0 & 0 & 0 & \cdots & b_n(y_t) 
\end{bmatrix}.
\]

\smallskip

We tested our \AI\ under different scenarios to determine whether or not the \AI\ could learn from its experiences to estimate the position of the mobile agent more accurately. To this end, we used the mean distance through out a game between the real position and the estimate as our statistic. We wanted to compare the performance of our \AI\ with adaptive memory with an \AI\ similar to Hladky's \cite{Hladky09}, which is not capable of machine learning, using the same statistic. We couldn't compare the precision of our estimates with estimates produced by a more recent \AI\ using reinforcement learning since those \AI's aren't actually trying to predict their opponent's location. We played multiple games using different strategies against various \AI's to collect data. For testing purposes we created multiple variations of the experiment.

\smallskip

First, we developed multiple strategies for the player. By a strategy, we mean a particular path from the starting location to one of the treasures that initially leads to victory. For example, a strategy could be to start by touching the top left parrots in order to gets the pirate to believe the player is going that way, then directly head to the right treasure by using the top lane.

\smallskip

 The first test set up was to use the exact same path for ten games in row. We've run this experiment with two different winning strategy. Note that intelligence with no learning abilities would lose those ten games. We did this to evaluate the capacity of our \AI\ with machine learning to adapt, to estimate the position of its opponent more precisely and finally to win. The second experiment consists of using a path for eight games then changing drastically to another winning strategy. We needed 4 winning precise path to run this experiment twice. This test was intended to verify the ability of our intelligence to recognize a change in strategy and adapt to the new tactic. Finally, we tried alternating between two winning strategies. We could thus examine how efficiently our \AI\ could learn two paths simultaneously. We have produced various graphics to demonstrate the evolution of our statistic throughout the experiment.

\smallskip

First, we present our results when we used the exact same tactic ten times in a row. We repeated this process with two different winning strategies. 

\smallskip
\begin{figure*}
\begin{center}
\includegraphics[width=16cm]{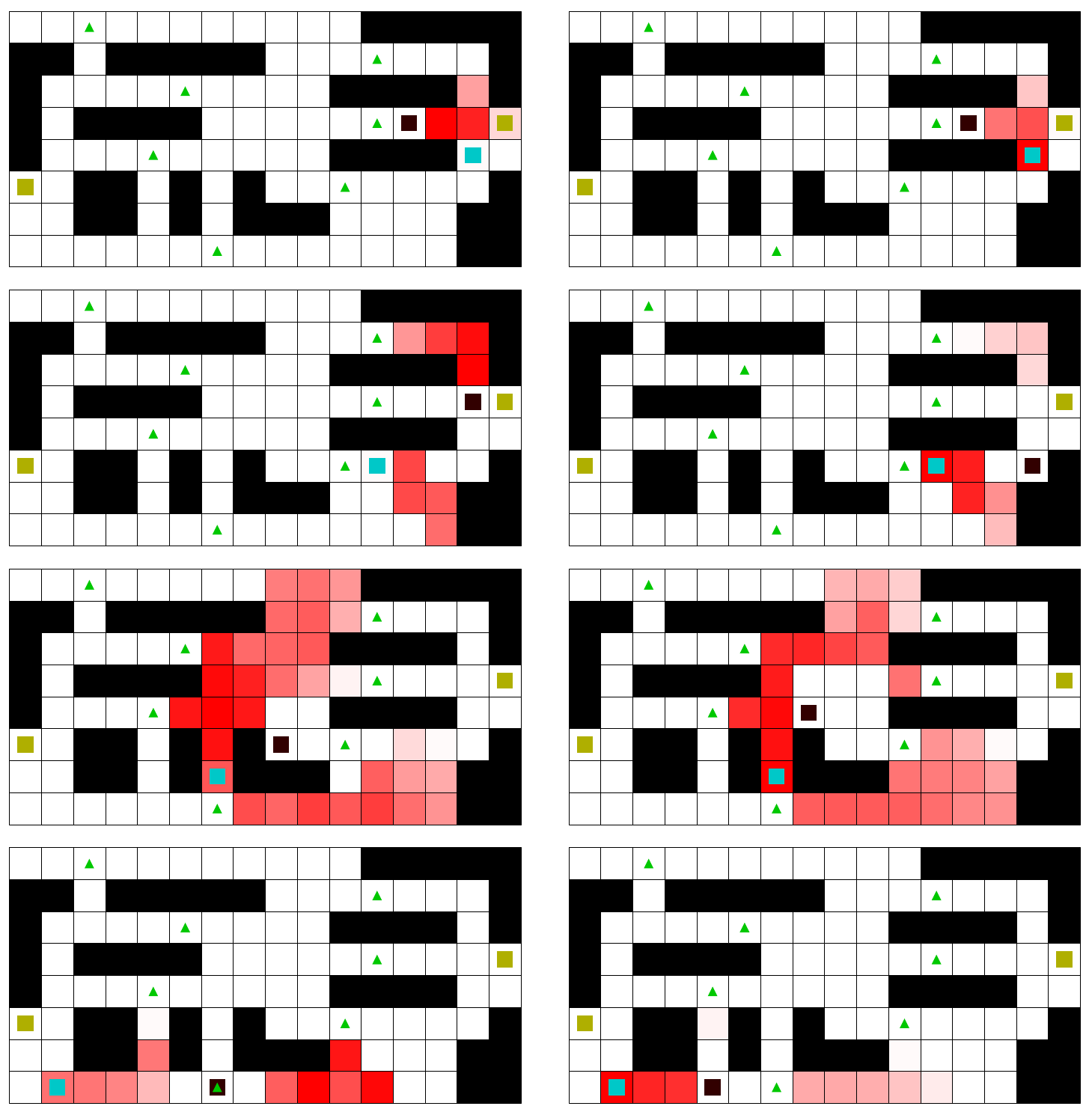}
\caption{These graphics represent the 12th, 16th, 24th and 30th turn of two games. The left column is the first game against a new player, while the graphics in the right column are from the tenth game against the same player. The blue square represents the human player, the intelligence's avatar is the dark brown square, the green triangles represent the parrots and objectives are represented by the golden squares. The player used the exact same path in all ten games.}
\label{heatmap1}
\end{center}
\end{figure*}

\begin{figure*}
\begin{center} 
\includegraphics[width=15cm]{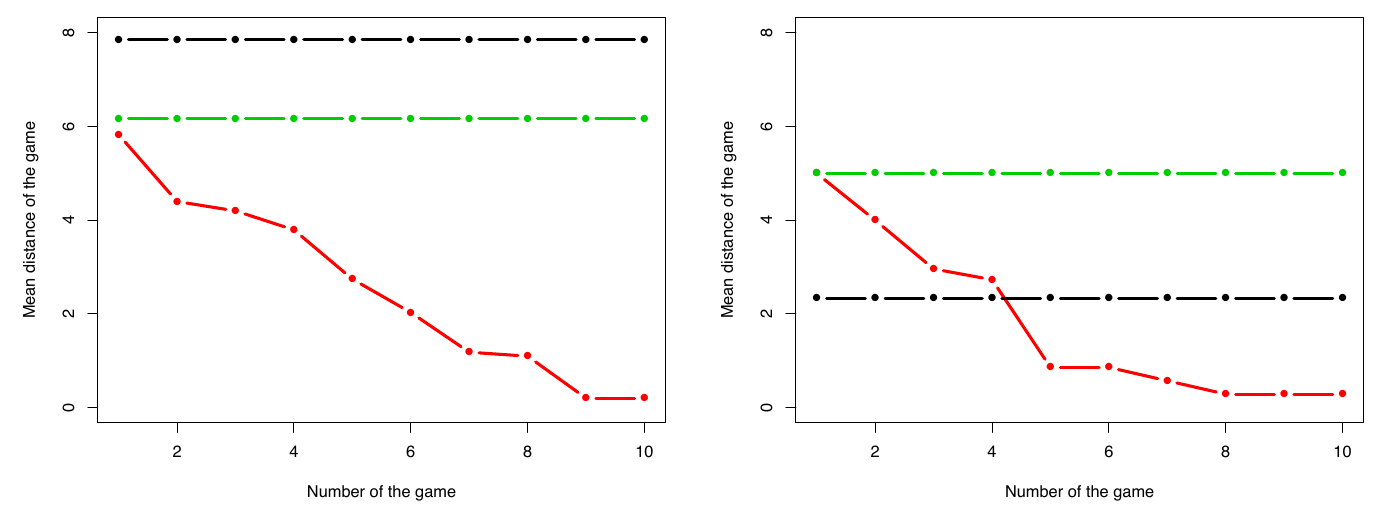}
\caption{Average distance for each game. Each figure represents a precise path. The green line represents an \AI\ with no machine learning and which used a uniformly distributed transition matrix. The black line represents the same \AI\ but this time one that used a transition matrix built using game experience, as Hladky did. Finally, the red line was obtained by playing against our \AI\ with adaptive memory.}
\label{comp1} 
\end{center}
\end{figure*}

We represent the real-time estimations and learning process elegantly via heat maps of probabilities. These maps are a representation of our experimental video game map in which we indicate the estimates performed by the artificial intelligence for each position at each turn. Rather than writing the exact probabilities we represented them using colors: black tiles are walls and do not count as possible positions; dark red tiles represent a high probability; white tiles represent the fact that the \AI's estimate for this position is 0. The player is identified as the blue square, and the intelligence's avatar is the dark brown square. The green triangles represent the parrots. Finally, objectives are represented by the golden squares.

\smallskip

The Figure \ref{heatmap1} contains heat maps from two different games at the exact same turns. The left  column represents the first game, and the right column represents the tenth, after the learning process. Not only can the graphics in Figure \ref{heatmap1} help us comprehend how this \AI\ works, that is how it sees the map, but they can also help us visualize the learning process. We can see the evolution of the state estimation obtained from the \AI\ with adaptive memory. In the left column, the player is frequently in the pale red tile, meaning the \AI\ does not consider this position to be highly probable. In contrast, the right column clearly shows that the player's real position is often painted in dark red. This illustrates that after multiple games, the \AI\ learned the path used by its opponent and now considers the agent's real position highly probable. We can also notice the difference in the path used by the intelligence to pursue its opponent. The \AI's avatar is much closer to the agent in the right column throughout the game.

\smallskip

We compare the performance of our \AI\ with machine learning to that of the \AI\ with no learning component using the statistics discussed earlier. Figure \ref{comp1} represents the evolution of the mean distance between the real position of the mobile agent and the estimation of the position from three different \AI's. We observe that for both of these strategies, the \AI\ with no learning component performs in exactly the same way in every single game. However, the mean distance between the mobile agent's position and the estimate from the \AI\ we designed decreases constantly. This indicates that our \AI's estimates are more precise as it learns from past games.

\smallskip


We finally tested that the statistic of interest was significantly smaller for the \AI\ we designed compared to Hladky's \AI. Our \AI's estimates were more precise in both of these situations ( t = -7.3049 with p-value = $6.225\times 10^{-09}$ and t-value = -3.2673 with p-value = 0.00619).

\smallskip

\begin{figure*}
\begin{center} 
\includegraphics[width=15cm]{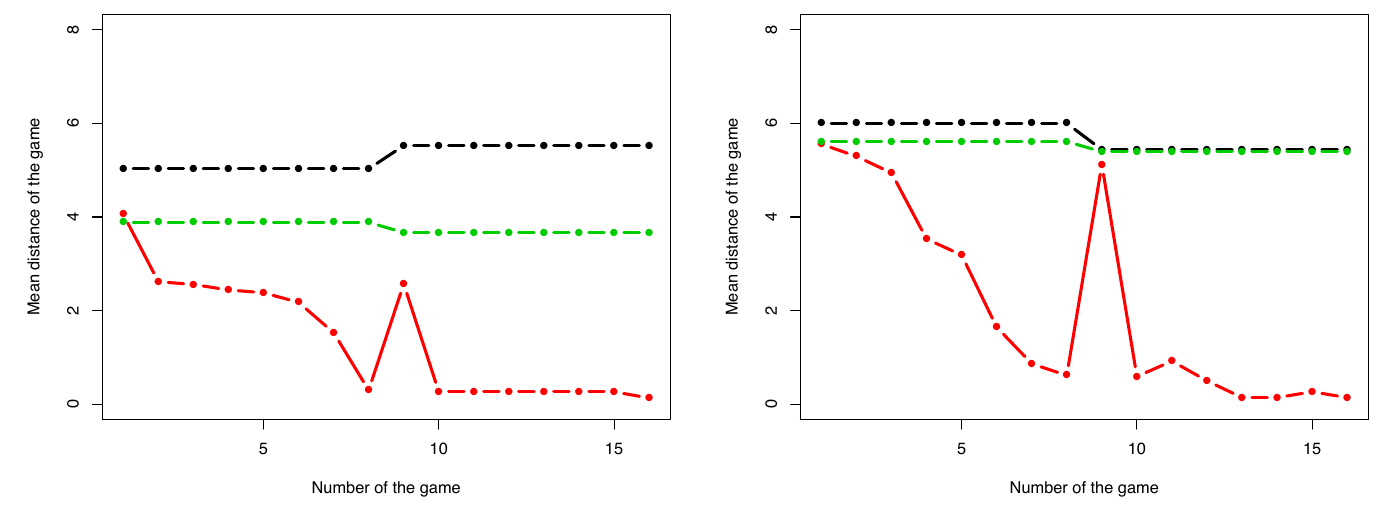}
\caption{Average distance for each game. Each figure represents a different run of the experiment. The green line represents an \AI\ with no machine learning and which used a uniformly distributed transition matrix. The black line represents the same \AI\ but this time one that used a transition matrix built using game experience, as Hladky did. Finally, the red line was obtained by playing against our \AI\ with adaptive memory.}
\label{comp2} 
\end{center}
\end{figure*}

It was important to verify how our \AI\ performed in other scenarios. Because a human player can adapt to multiple strategies, we had to be sure that our intelligence would adapt to an abrupt change in tactics. That is why we played eight games using the same strategy, then completely changed our strategy to a new one for the eight following games. Figure \ref{comp2} contains the evolution of the test statistics we designed. It illustrates that the \AI\ with adaptive memory produces superior estimates after the learning process. Even though the change in strategy surprises the \AI\ at the ninth confrontation causing poor performance, we can clearly see that our \AI\ adapts quickly to this new tactic.  

\smallskip

Note that the \AI\ we developed had more precise estimates after the learning process. The difference was statistically significant for the eighth game (t-value = -6.3822 with p-value = $9.97 \times 10^{-08}$ and t-value = -7.2901 with p-value = $4.362 \times 10^{-08}$) and for the 16th game (t-value = -5.0932 with p-value = $2.241 \times 10^{-05}$ and t-value = -8.0063 with p-value = $1.387 \times 10^{-09}$). Finally, we did not want the learning process to become a handicap. It was important for us that to ensure that after learning a certain strategy, our \AI\ would not produce poor estimates for every other strategy. That is why we also compared our test statistic right after the change of strategy. We can confirm that even after learning a tactic, our \AI's estimates are as precise as an \AI\ with no machine learning (t-value = -1.1967 with p-value = 0.2386 and t-value = -0.2683 with p-value = 0.7894). 
 
\smallskip

\begin{figure*}
\label{comparaison} 
\begin{center}
\includegraphics[width=15cm]{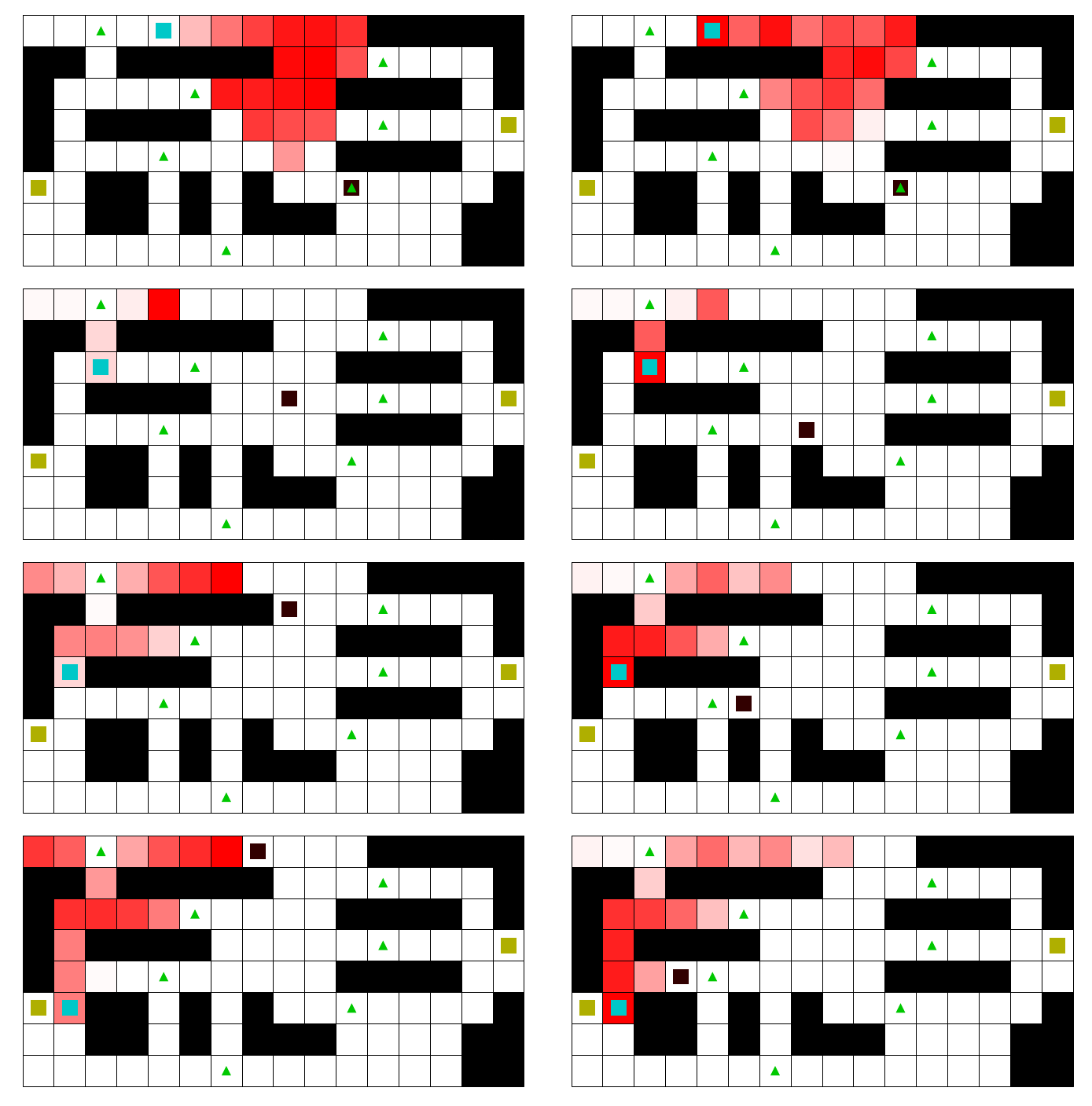}
\caption{These graphics represent the 6th, 10th, 12th and 14th turn of two games. The left column is the first game against a new player, while the right column represents the eleventh game against the same player. The blue square represents the human player, the intelligence's avatar is the dark brown square, the green triangles represent the parrots and objectives are represented by the golden squares. The player alternated between two strategies.}
\label{heatmap2}
\end{center}
\end{figure*}

Our last experiment was set up to check how our \AI\ would perform against an opponent that alternated between two initially winning strategies. We wanted to verify our \AI's ability to learn more than one strategy simultaneously. To demonstrate the learning process we present another set of heat maps, shown in Figure \ref{heatmap2}. 

\smallskip

Once again, we can notice a substantial difference in the shade of red of the player's tile, as shown in Figure \ref{heatmap2}. We can also easily observe a difference in the path that the \AI's avatar used. This time it was also much closer to the human player in the 11th game than in the first game.

\smallskip
\begin{figure*}
\begin{center} 
\includegraphics[width=15cm]{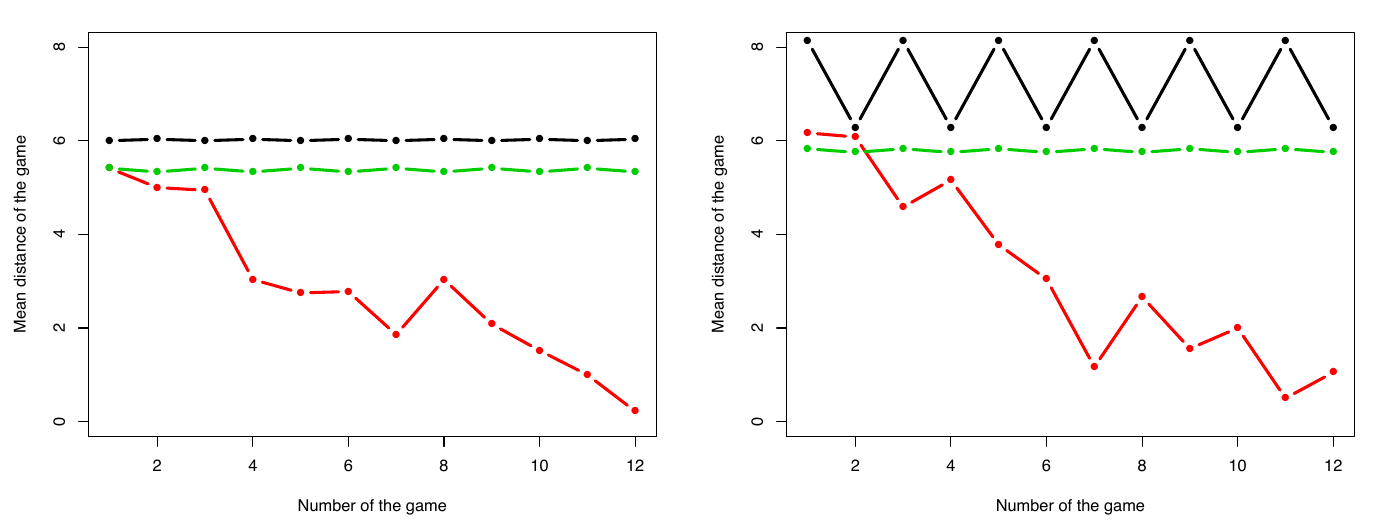}
\caption{Average distance for each game.  Each figure represents a different run of the experiment. The green line represents an \AI\ with no machine learning and which used a uniformly distributed transition matrix. The black line represents the same \AI\ but this time one that used a transition matrix built using game experience, as Hladky did. Finally, the red line was obtained by playing against our \AI\ with adaptive memory.}
\label{comp3}
\end{center}
\end{figure*}

Lastly, we graphically illustrate the test statistic for this experiment. Figure \ref{comp3} indicated that our \AI\ slowly improved its estimates. The process is a bit more unstable and slower than the process with only one strategy, but it is still highly effective. We can see that our \AI\ is able to learn at least two strategies simultaneously.

\smallskip

We now verify that the intelligence with adaptive memory produced more precise estimations after the learning process took place. Since it is constantly switching between two strategies, we examined the 11th and the 12th matches. The mean distance between the estimate and the real position of our \textit{bot} was smaller both for the 11th game ( t-value = -5.3526 with p-value = $6.451 \times 10^{-06}$ and t-value = -6.8703 with p-value = $1.983 \times 10^{-08}$) and the 12th game ( t-value = -7.8525 with p-value = $1.783 \times 10^{-09}$ and t-value = -6.2715 with p-value = $2.553 \times 10^{-07}$).

\section{Conclusion}

We have developed a way to represent the problem of tracking a mobile agent mathematically, specifically, through the use of a Hidden Markov Model. After finding a way to efficiently program the \textit{HMM}'s algorithms, we built artificial intelligence that could use this model to accurately estimate the unknown position of a mobile agent in a restricted area. 

\smallskip

We programmed our own video game for testing purposes, and constructed test statistics and multiple graphical tools to visualize the effect of our work. We found that \AI\ demonstrated very strong results; the real-time estimation procedure combined with the learning process made it difficult for human players to win after a number of rounds. Furthermore, the processing time involved in the real-time estimation component was efficient to the point of not being noticeable, which affords it the potential to be implemented in numerous applications. Finally, the adaptive memory also produced very strong results. Although it requires more time to adjust the various settings precisely, our \AI\ can be a powerful tool in many applications, especially video games applications. 

\smallskip

Much work remains before statistics and probability models can be used efficiently in the construction of artificial intelligence, but we believe that our work described above is evidence that this kind of interdisciplinary research has great potential. We do hope that statistics become part of the culture of development of new artificial intelligence, and that our work has contributed to that trend. 

\section*{Acknowledgments}
The authors personally thank Louis-Alexandre Valli\`{e}res-Lavoie and Guillaume Racicot for their impressive work on the program. The authors also gratefully acknowledge the financial support from the NSERC and the FRQNT. Finally, the authors would like to thank Sorana Froda for her important contribution and Ali Al-Aradi for his most helpful suggestions.


\bibliographystyle{./IEEEtran}
\bibliography{./IEEEabrv,./mybibfile}

\end{document}